\documentclass[conference]{IEEEtran}

\usepackage{amsmath,amsfonts}
\usepackage{algorithmic}
\usepackage{array}
\usepackage[caption=false,font=normalsize,labelfont=sf,textfont=sf]{subfig}
\usepackage{textcomp}
\usepackage{stfloats}
\usepackage{url}
\usepackage{verbatim}
\def\BibTeX{{\rm B\kern-.05em{\sc i\kern-.025em b}\kern-.08em
    T\kern-.1667em\lower.7ex\hbox{E}\kern-.125emX}}
\usepackage{balance}
\usepackage{moreverb,url}
\usepackage{graphicx}
\usepackage{xcolor}
\usepackage[colorlinks,bookmarksopen,bookmarksnumbered,citecolor=red,urlcolor=red]{hyperref}
\usepackage{lipsum}
\usepackage{booktabs}

\usepackage{soul}
\soulregister\ref7
\soulregister\citep7
\usepackage{float}
\usepackage{graphicx}
\usepackage{enumitem}
\usepackage{footmisc}

\usepackage{eso-pic}
\newcommand\AtPageUpperMyright[1]{\AtPageUpperLeft{%
 \put(\LenToUnit{0.17\paperwidth},\LenToUnit{-2cm}){%
     \parbox{0.9\textwidth}{\raggedleft\fontsize{8}{11}\selectfont #1}}%
 }}%
 \newcommand{\conf}[1]{%
\AddToShipoutPictureBG*{%
\AtPageUpperMyright{#1}
}}

 \newcommand\blfootnote[1]{%
  \begingroup
  \renewcommand\thefootnote{}\footnote{#1}%
  \addtocounter{footnote}{-1}%
  \endgroup
}

\begin{document}
\title{\vspace*{1cm} A surgical dataset from the da Vinci Research Kit for task automation and recognition\\}

\author{\IEEEauthorblockN{1\textsuperscript{st} Irene Rivas-Blanco}
\IEEEauthorblockA{\textit{Dept. Systems Engineering and Automation}\\
\textit{University of Malaga}\\
Malaga, Spain \\
irivas@uma.es}
\and
\IEEEauthorblockN{2\textsuperscript{nd} Carlos J. Pérez-del-Pulgar}
\IEEEauthorblockA{\textit{Dept. Systems Engineering and Automation}\\
\textit{University of Malaga}\\
Malaga, Spain \\
carlosperez@uma.es}
\and
\IEEEauthorblockN{3\textsuperscript{rd} Andrea Mariani}
\IEEEauthorblockA{\textit{The BioRobotics Institute} \\
\textit{Dept. Excellence in Robotics \& AI}\\
\textit{Scuola Superiore Sant'Anna}\\
Pisa, Italy \\
andrea.mariani@santannapisa.it}
\and
\IEEEauthorblockN{4\textsuperscript{th} Giuseppe Tortora}
\IEEEauthorblockA{\textit{Smart Medical Theatre Lab} \\
\textit{ABzero srl}\\
Pisa, Italy \\
giuseppe.tortora@santannapisa.it}
\and
\IEEEauthorblockN{5\textsuperscript{th} Antonio J. Reina}
\IEEEauthorblockA{\textit{Dept. Systems Engineering and Automation}\\
\textit{University of Malaga}\\
Malaga, Spain \\
ajreina@uma.es}
}

\maketitle
\conf{\textit{  Proc. of the International Conference on Electrical, Computer, Communications and Mechatronics Engineering (ICECCME 2023) \\ 
19-20 July 2023, Tenerife, Canary Islands, Spain}}

\begin{abstract}
The use of large datasets is essential in surgical robotics to advance in the field of recognition and automation of surgical tasks. Furthermore, public datasets allow the comparison of different algorithms and methods to evaluate their performance. The objective of this work is to provide a complete dataset of three common training surgical tasks performed with the da Vinci Research Kit (dVRK). The dataset contains a total of 206 trials performed by twelve different subjects. For each trial, the dataset includes 154 kinematic variables from the dVRK (both master and slave sides) together with the associated video recorded with a camera. All the data has been carefully timestamped and provided in a readable csv format. A MATLAB interface integrated with ROS for using and replicating the data is also provided.   
\end{abstract}

\begin{IEEEkeywords}
Dataset, da Vinci Research Kit, automation, surgical robotics
\end{IEEEkeywords}

\blfootnote{ This work was supported by the Spanish Ministry of Science and Innovation, under grant number PID2021-125050OA-I00.}

\section{Introduction}
Surgical Data Science (SDS) is emerging as a new knowledge domain in healthcare. In the field of surgery, it can provide many advances in virtual coaching, surgeon skill evaluation and complex tasks learning from surgical robotic systems \cite{Vedula2020SurgicalDomain}, as well as in the gesture recognition domain \cite{Perez-del-Pulgar2019}, \cite{Ahmidi2017ASurgery}. Surgical scene understanding has become an essential task for developing intelligent systems able to collaborate with surgeons during a real intervention \cite{Saras}. The development of large datasets related to the execution of surgical tasks using robotic systems would support these advances, providing detailed information of the surgeon movements, in terms of both kinematics and dynamics data, as well as video recordings. Moreover, public datasets allows to compare the performance of different algorithms proposed in the literature. 

Rivas-Blanco et al. \cite{Rivas-Blanco2021ASurgery} provides a list of 13 publicly available datasets in the surgical domain. Most datasets include video data, but only two of them incorporate kinematic data, which provides a big amount of useful information for analyzing metrics related to the motion of the tools. Kinematic data is recorded from a da Vinci Research Kit (dVRK), a research platform based on the first-generation commercial da Vinci Surgical System (by Intuitive Surgical, Inc., Sunnyvale, CA). This platform has a software package that provides kinematics and dynamics data of the Master Tool and the Patient Side Manipulators.

The JIGSAWS dataset, described in \cite{Gao2014JHU-ISIModeling} is the most known dataset in surgical robotics. This dataset includes 76-dimensional kinematic data along with the video data for 101 trials of three elementary surgical tasks (suturing, knot-tying, and needle-passing) performed by 6 surgeons using the dVRK. 
On the other hand, the UCL dVRK dataset \cite{SL18} contains 14 videos using the dVRK on five different kinds of animal tissue. For each video frame, an associated image of the virtual tools is produced using a dVRK simulator. 

The objective of this work is to extend the publicly available datasets related to surgical robotics. We present the Robotic Surgical Manuevers (ROSMA) dataset, a large dataset collected using the dVRK, in collaboration between the University of Malaga (Spain) and The Biorobotics Insitute of the Scuola Superiore Sant'Anna (Italy), under a TERRINet (The European Robotics Research Infrastructure Network) project.
This dataset contains 36 kinematic variables, divided into 154-dimensional data, recorded at 50 Hz for 206 trials of three common training surgical tasks. This data is complemented with the video recordings collected at 15 frames per second with 1024 x 768 pixel resolution. Moreover, we provide a task evaluation based on time and task-specific errors, a synchronization data file between data and videos, the transformation matrix between the camera and the Patient Side Manipulators, and a questionnaire with personal data of the subjects (gender, age, dominant hand) and previous experience using teleoperated systems and visuo-motor skills (sport and musical instruments). We expect that the data provided in this work would be helpful for future research on automation and recognition of surgical tasks. The simplicity of the tasks recorded facilitates the performance of dry lab experiments based on these data.
In summary, the main contributions of this letter are:

\begin{enumerate}
    \item To provide a publicly available large dataset of surgical robotic tasks collected with the da Vinci Research Kit, including kinematic and video data. 
    \item To complete the dataset with the camera projection matrixes that relates the 3D coordinates of the tool tips to 2D image coordinates. 
    \item To facilitate the usage of the data, it is provided a MATLAB software that gives the option of reproducing the data of the experiments in ROS topics.
\end{enumerate}

\section{System description} 

\begin{figure*}[!ht]
    \centering
  \subfloat[Slave manipulators\label{fig:platform:a}]{%
       \includegraphics[width=0.40\linewidth]{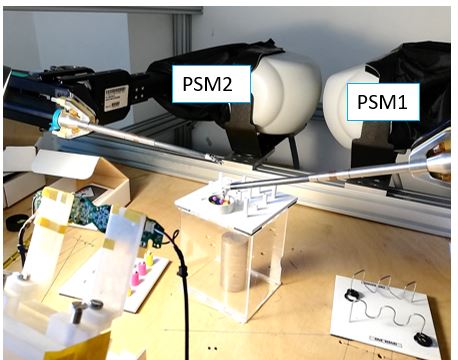}}
    \hfill
  \subfloat[Master console\label{fig:platform:b}]{%
        \includegraphics[width=0.42\linewidth]{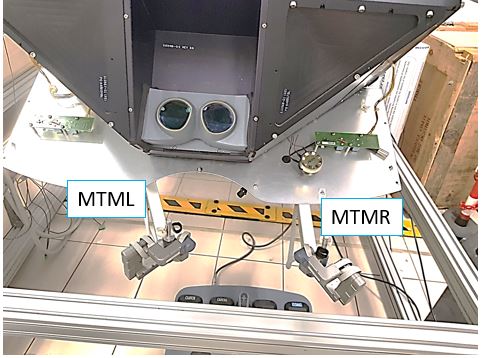}}
  \caption{da Vinci Research Kit platform available at The Biorobotics Institute of Scuola Superiore Sant'Anna (Pisa, Italy)}
  \label{fig1} 
\end{figure*}

\begin{figure*}[!ht]
    \centering
    \subfloat[Patient side kinematics\label{fig:kinematics:PSM}]{
        \includegraphics[width=0.40\textwidth]{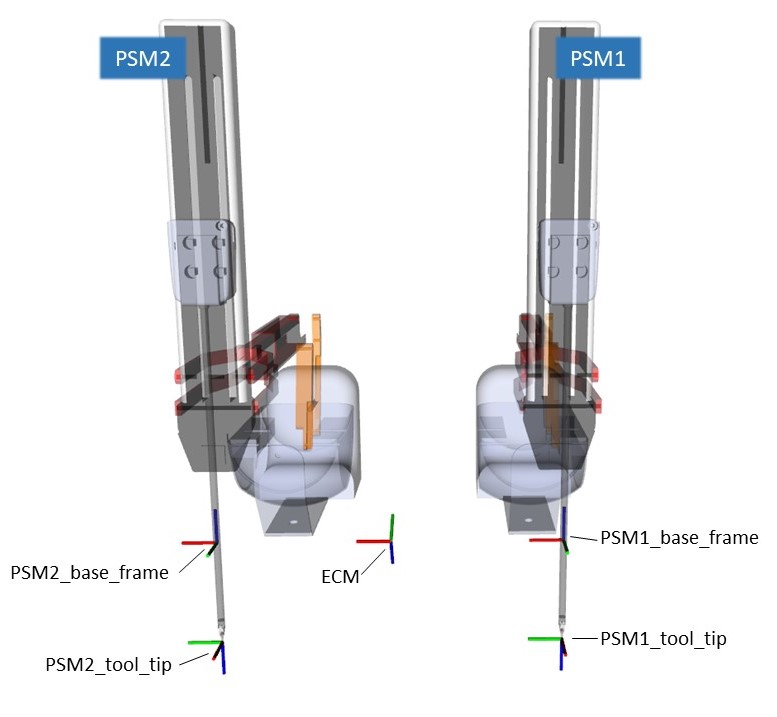}}
    \hfill
    \subfloat[Surgeon side kinematics\label{fig:kinematics:MTM}]{
        \includegraphics[width=0.40\textwidth]{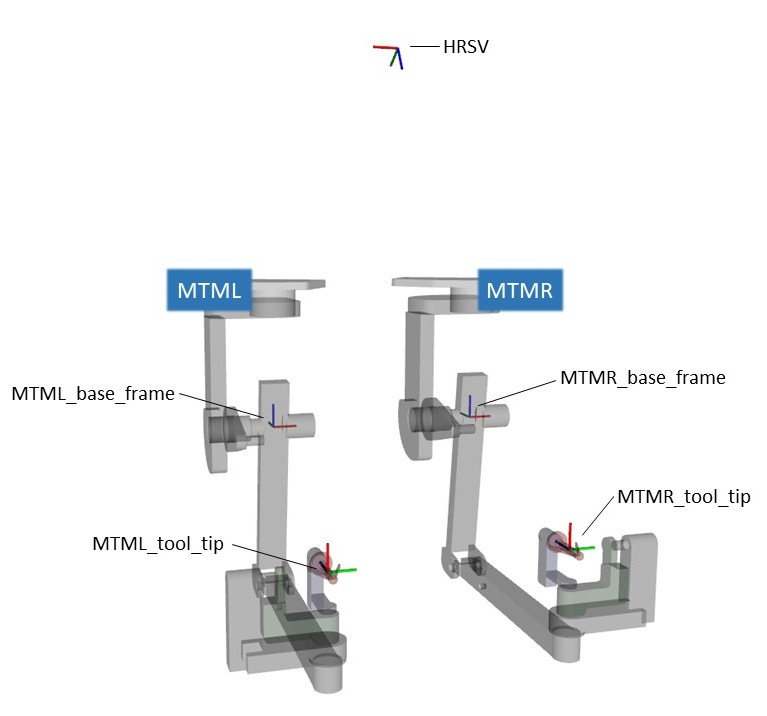}}
    \caption{Patient and surgeon side kinematics. Kinematics of each PSM is defined with respect to the common frame ECM, while the MTMs are described with respect to frame HRSV.}
\end{figure*}

\begin{table*}[t!]
    \small\sf 
    \caption{Protocol for each task of the ROSMA dataset}
    \begin{tabular}{p{0.11\textwidth}p{0.27\textwidth}p{0.27\textwidth}p{0.27\textwidth}}
    \toprule
        &Post and sleeve&Pea on a peg&Wire chaser\\
    \midrule
        \texttt{Goal}
            & To move the colored sleeves from side to side of the board.
            & To put the beads on the 14 pegs of the board.  
            & To move a ring from one side to the other side of the board. \\
        \texttt{Starting position}  
            & The board is placed with the peg rows in a vertical position (from left to right: 4-2-2-4). The six sleeves are positioned over the 6 pegs on one of the sides of the board.
             & All beads are on the cup. 
            & The board is positioned with the text “one hand” in front. The three rings are at the right side of the board.\\ 
        \texttt{Procedure} 
            & The subject has to take a sleeve with one hand, pass it to the other hand, and place it over a peg on the opposite side of the board. If a sleeve is dropped, it is considered a penalty and it cannot be taken back. 
            & The subject has to take the beads one by one out of the cup and place them on top of the pegs. For the trials performed with the right hand, the beads are placed on the right side of the board, and vice versa. If a bead is dropped, it is considered a penalty and it cannot be taken back. 
            & The subject has to pick one of the rings and pass it through the wire to the other side of the board. The subjects must use only hand to move the ring, but they are allowed to help themselves with the other hand if needed. If the ring is dropped, it is considered a penalty but it must be taken back to complete the task.\\
        \texttt{Repetitions} 
            & Six trials: three from right to left, and other three from left to right.
            & Six trials: three placing the beads on the pegs of the right side of the board, and other three on the left side.
            & Six trials: three moving the rings from right to left, and other three from left to right.\\
        \texttt{Penalty} 
            & 15 penalty points if a sleeve is dropped.
            & 15 penalty points when a bead is dropped.
            & 10 penalty points when the ring is dropped.\\
        \texttt{Score} 
            & Time in seconds + penalty points.
            & Time in seconds + penalty points.
            & Time in seconds + penalty points.\\
    \bottomrule
    \end{tabular}
    * A trial is the data corresponding with the performance of one subject one instance of a specific task.
    \label{table:protocol}
\end{table*}

\begin{figure*}[!ht]
   \centering
    \subfloat[Post and sleeve\label{fig:task:PS}]{
        \includegraphics[width=0.33\textwidth]{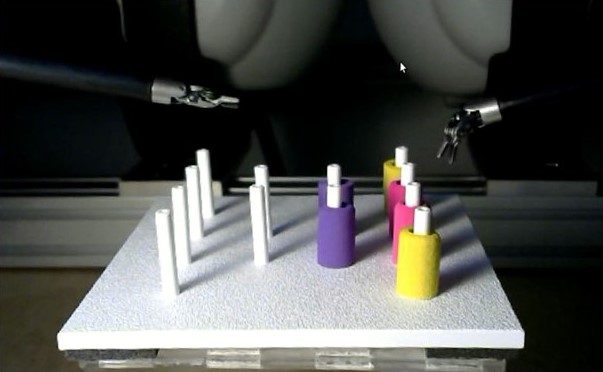}}
    \subfloat[Pea on a peg\label{fig:task:PP}]{
        \includegraphics[width=0.33\textwidth]{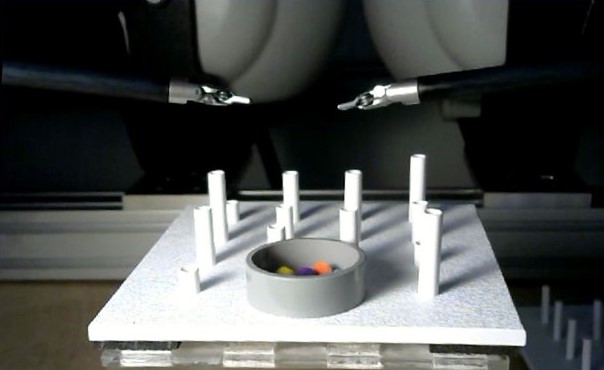}}
    \subfloat[Wire Chaser\label{fig:task:WC}]{
        \includegraphics[width=0.33\textwidth]{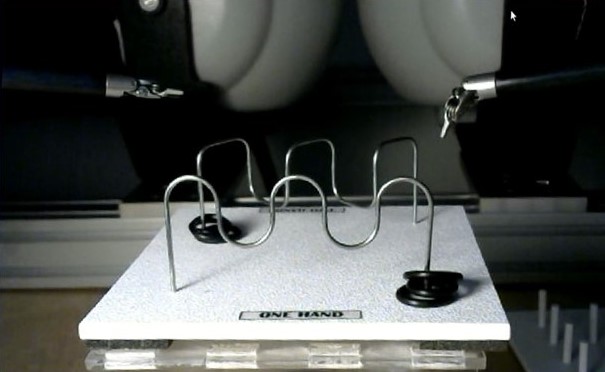}}
    \caption{Snapshot of the three tasks in the ROSMA dataset at the starting position.}
\end{figure*}

\begin{table*}[!ht]
    \small\sf\centering
    \caption{Dataset summary: number of trials of each subject and exercise in the ROSMA dataset.}
    \begin{tabular}{llllllllllllll}
    \toprule
        &X01&X02&X03&X04&X05&X06&X07&X08&X09&X10&X11&X12&Total\\
    \midrule
        \texttt{Post and sleeve}&6&6&5&5&5&6&6&6&4&6&6&4&65\\
        \texttt{Pea on a peg}&6&6&6&6&6&6&6&6&5&6&6&6&71\\
        \texttt{Wire chaser}&6&6&5&6&6&6&6&6&6&5&6&6&70\\
        \texttt{No. Trials per subject}&18&18&16&17&17&18&18&18&16&17&17&16&206\\
    \bottomrule
    \end{tabular}
    \label{table:summary}
\end{table*}

The dVRK, supported by the Intuitive Foundation (Sunnyvale, CA), arose as a community effort to support research in the field of telerobotic surgery \cite{Kazanzides2014AnSystem}. This platform is made up of hardware of the first-generation da Vinci system along with motor controllers and a software framework integrated with the Robot Operating System (ROS) \cite{Chen2017SoftwareKit}. There are over thirty dVRK platforms distributed in ten countries around the world. 
The Biorobotics Institute of the Scuola Superiore Sant'Anna (Pisa, Itally) has a dVRK with two Patient Side Manipulators (PSM), labelled as PSM1 and PSM2 (Figure \ref{fig:platform:a}), and a master console consisting of two Master Tool Manipulators (MTM), labelled as MTML and MTMR (Figure \ref{fig:platform:b}). MTMR controls PSM1, while MTML controls PSM2. For the experiments described in this paper, the stereo vision is provided using two commercial webcams, as the dVRK used for the experiments was not equipped with the endoscopic camera manipulator. Each PSM has 6 joints following the kinematics described in \cite{Fontanelli2017ModellingArms}, and an additional degree of freedom for opening and closing the gripper. The tip of the instrument moves about a remote center of motion, where the origin of the base frame of each manipulator is set. The motion of each manipulator is described by its corresponding \textit{tool\_tip} frame with respect to the home position of the PSM.
The MTMs used to remotely teleoperate the PSMs have 7-DOF, plus the opening and closing of the instrument. The base frames of each manipulator are related through the common frame \textit{HRSV}, as show in Figure \ref{fig:kinematics:MTM}. The transformation between the base frames and the common one in both sides of the dVRK is described in the json configuration file that can be found in the ROSMA Github repository\footnote{\label{git}\url{https://github.com/SurgicalRoboticsUMA/rosma_dataset}}.

The ROSMA dataset contains the performance of three tasks of the Skill Building Task Set (from 3-D Techical Services, Franklin, OH): post and sleeve (Figure \ref{fig:task:PS}), pea on a peg (Figure \ref{fig:task:PP}), and Wire Chaser (Figure \ref{fig:task:WC}). These training platforms for clinical skill development provide challenges that require motions and skills used in laparoscopic surgery, such as hand-eye coordination, bimanual dexterity, depth perception or interaction between dominant and non-dominant hand. These exercises were born for clinical skill acquisition, but they are also commonly used in surgical research for different applications, such as clinical skills acquisition \cite{Hardon2018Force-basedTraining} or testing of new devices \cite{Velasquez2016PreliminarySurgery}. The protocol for each exercise is described in Table \ref{table:protocol}. 

\section{Dataset structure}

\begin{table*}[!ht]
    \small\sf\centering
    \caption{Structure of the columns of the data files.}
    \begin{tabular}{lllll}
    \toprule
        Indices& Number &Label & Unit & ROS publisher\\
    \midrule
        \texttt{2-4}& 3& MTML\_position\_\textless x,y,z\textgreater & m & \textit{/dvrk/MTML/position\_cartesian\_current/position} \\
        \texttt{5-8}&4& MTML\_orientation\_\textless x,y,z,w\textgreater & radians & \textit{/dvrk/MTML/position\_cartesian\_current/orientation} \\
        \texttt{9-11}& 3&MTML\_velocity\_linear\_\textless x,y,z\textgreater & m/s & \textit{/dvrk/MTML/twist\_body\_current/linear}  \\
        \texttt{12-14}&3& MTML\_velocity\_angular\_\textless x,y,z\textgreater & radians/s & \textit{/dvrk/MTML/twist\_body\_current/angular} \\
        \texttt{15-17}&3& MTML\_wrench\_force\_\textless x,y,z\textgreater & N & \textit{/dvrk/MTML/wrench\_body\_current/force}  \\
        \texttt{18-20}&3& MTML\_wrench\_torque\_\textless x,y,z\textgreater & N·m & \textit{/dvrk/MTML/wrench\_body\_current/torque} \\
        \texttt{21-27}&7& MTML\_joint\_position\_\textless 1,2,3,4,5,6,7\textgreater & radians & \textit{/dvrk/MTML/state\_joint\_current/position}  \\
        \texttt{28-34}&7& MTML\_joint\_velocity\_\textless 1,2,3,4,5,6,7\textgreater & radians/s & \textit{/dvrk/MTML/state\_joint\_current/velocity} \\
        \texttt{35-41}&7& MTML\_joint\_effort\_ \textless 1,2,3,4,5,6,7\textgreater & N & \textit{/dvrk/MTML/state\_joint\_current/effort}  \\
        \texttt{42-44}&3& MTMR\_position\_\textless x,y,z\textgreater & m & \textit{/dvrk/MTMR/position\_cartesian\_current/position} \\
        \texttt{45-48}&4& MTMR\_orientation\_\textless x,y,z,w\textgreater & m & \textit{/dvrk/MTMR/position\_cartesian\_current/orientation} \\
        \texttt{49-51}&3& MTMR\_velocity\_linear\_\textless x,y,z\textgreater & m/s & \textit{/dvrk/MTMR/twist\_body\_current/linear}  \\
        \texttt{52-54}&3& MTMR\_velocity\_angular\_\textless x,y,z\textgreater & radians/s & \textit{/dvrk/MTMR/twist\_body\_current/angular} \\
        \texttt{55-57}&3& MTMR\_wrench\_force\_\textless x,y,z\textgreater & N & \textit{/dvrk/MTMR/wrench\_body\_current/force}  \\
        \texttt{58-60}&3& MTMR\_wrench\_torque\_\textless x,y,z\textgreater & N·m & \textit{/dvrk/MTMR/wrench\_body\_current/torque} \\
        \texttt{61-67}&7& MTMR\_joint\_position\_ \textless 1,2,3,4,5,6,7\textgreater & radians & \textit{/dvrk/MTMR/state\_joint\_current/position} \\
        \texttt{68-74}&7& MTMR\_joint\_velocity\_\textless 1,2,3,4,5,6,7\textgreater & radians/s & \textit{/dvrk/MTMR/state\_joint\_current/velocity} \\
        \texttt{75-81}&7& MTMR\_joint\_effort\_\textless 1,2,3,4,5,6,7\textgreater & N & \textit{/dvrk/MTMR/state\_joint\_current/effort} \\   
        \texttt{82-84}&3& PSM1\_position\_\textless x,y,z\textgreater & m & \textit{/dvrk/PSM1/position\_cartesian\_current/position} \\
        \texttt{85-88}&4& PSM1\_orientation\_\textless x,y,z,w \textgreater & radians & \textit{/dvrk/PSM1/position\_cartesian\_current/orientation}\\
        \texttt{89-91}&3& PSM1\_velocity\_linear\_\textless x,y,z\textgreater & m/s & \textit{/dvrk/PSM1/twist\_body\_current/linear}  \\
        \texttt{92-94}&3& PSM1\_velocity\_angular\_\textless x,y,z\textgreater & radians/s & \textit{/dvrk/PSM1/twist\_body\_current/angular} \\
        \texttt{95-97}&3& PSM1\_wrench\_force\_\textless x,y,z\textgreater & N & \textit{/dvrk/PSM1/wrench\_body\_current/force}  \\
        \texttt{98-100}&3& PSM1\_wrench\_torque\_\textless x,y,z\textgreater & N·m & \textit{/dvrk/PSM1/wrench\_body\_current/torque} \\
        \texttt{101-106}&6& PSM1\_joint\_position\_\textless 1,2,3,4,5,6\textgreater & radians & \textit{/dvrk/PSM1/state\_joint\_current/position} \\
        \texttt{107-112}&6& PSM1\_joint\_velocity\_\textless 1,2,3,4,5,6\textgreater & radians/s & \textit{/dvrk/PSM1/state\_joint\_current/velocity} \\
        \texttt{113-118}&6& PSM1\_joint\_effort\_\textless 1,2,3,4,5,6\textgreater & N & \textit{/dvrk/PSM1/state\_joint\_current/effort} \\
        \texttt{118-120}&3& PSM2\_position\_\textless x,y,z\textgreater & m & \textit{/dvrk/PSM2/position\_cartesian\_current/position} \\
        \texttt{120-124}&4& PSM2\_orientation\_\textless x,y,z,w\textgreater & radians & \textit{/dvrk/PSM2/position\_cartesian\_current/orientation} \\
        \texttt{125-128}&3& PSM2\_velocity\_linear\_\textless x,y,z\textgreater & m/s & \textit{/dvrk/PSM2/twist\_body\_current/linear}\\
        \texttt{129-131}&3& PSM2\_velocity\_angular\_ \textless x,y,z\textgreater & radians/s & \textit{/dvrk/PSM2/twist\_body\_current/angular} \\
        \texttt{132-134}&3& PSM2\_wrench\_force\_\textless x,y,z\textgreater & N &\textit{/dvrk/PSM2/wrench\_body\_current/force} \\
        \texttt{135-137}&3& PSM2\_wrench\_torque\_\textless x,y,z\textgreater & N·m & \textit{/dvrk/PSM2/wrench\_body\_current/torque} \\
        \texttt{138-143}&6& PSM2\_joint\_position\_\textless 1,2,3,4,5,6\textgreater & radians & \textit{/dvrk/PSM2/state\_joint\_current/position} \\
        \texttt{144-149}&6& PSM2\_joint\_velocity\_\textless 1,2,3,4,5,6\textgreater & radians/s & \textit{/dvrk/PSM2/state\_joint\_current/velocity} \\
        \texttt{150-155}&6& PSM2\_joint\_effort\_\textless 1,2,3,4,5,6\textgreater & N & \textit{/dvrk/PSM2/state\_joint\_current/effort} \\     
    \bottomrule
    \end{tabular}
    \label{table:columns}
\end{table*}

The ROSMA dataset is divided into three training tasks performed by twelve subjects. The experiments were carried out in accordance with the recommendations of our institution with written informed consent from the subjects in accordance with the declaration of Helsinki. Before starting the experiment, each subject was taught about the goal of the exercises and the error metrics. The overall length of the data recorded is 8 hours, 19 minutes and 40 seconds, and the total amount of kinematic data is around 1.5 millions for each parameter. The dataset contains 415 files divided as follows: 206 data files in csv (comma-separated values) format; 206 video data files in mp4 format; a file in cvs format with the exercises evaluation, named \textit{'scores.csv'}; a file in txt format with the synchronization data between the csv files and the video files, named \textit{'synchronizationData.txt'};and a file, also in csv format, with the answers of the personal questionnaire, named \textit{'User questionnaire - dvrk Dataset Experiment.csv'}. 

The name of the data and video files follows the hierarchy: \textless User\_Id\textgreater\_\textless Task\_name\textgreater\_\textless Trial\_number\textgreater. The description of each of these fields is as follows: 
\begin{itemize}
    \item \textless User\_Id\textgreater: it provides a unique identifier for each user, and ranges from 'X01' to 'X12'.
    \item \textless Task\_name\textgreater: it may be one of the following labels, depending on the task being performed: 'Pea\_on\_a\_Peg', 'Post\_and\_Sleeve', or 'Wire\_Chaser'
    \item \textless Trial\_number\textgreater: it is the repetition number of the user in the current task, ranging from '01' to '06'. 
\end{itemize}
For example, the file name 'X03\_Pea\_on\_a\_Peg\_04' corresponds to the fourth trial of the user 'X03' performing the task Pea on a Peg. 

The summary of the number of trials performed by each user and task is shown in Table \ref{table:summary}. Each user performed a total of six trials per task, but during the post-processing of the data, the authors found recording errors in some of them. That is the reason why there are some users with fewer trials in certain tasks. The full dataset is available for download at the Zenodo website \cite{Rivas-Blanco2020TrainingKit}.  

\subsection{Data files}
Data files are in csv format and contain 155 columns: the first column, labelled as 'Date' has the timestamp of each set of measures, and the other 154 columns have the kinematic data of the patient side manipulators (PSMs) and the master side manipulators (MSMs). The structure of these 154 columns is described in Table \ref{table:columns}, which shows the column indices for each kinematic motion, the number of columns, the descriptive label of each variable, the data units, and the ROS publishers of the data. The descriptive labels of the columns has the following format: \textless component\_name\textgreater \_\textless kinematic\_motion\textgreater\_\textless variable\textgreater. 
The timestamp values have a precision of milliseconds, and are expressed in the format: \textit{Year-Month-Day.Hour:Minutes:Seconds.Milliseconds}. As data has been recorded at 50 samples per seconds, the time step between rows is 20 ms. 

\subsection{Video files}
Images have been recorded from one of the two commercial webcams used during the experiments to achieve the stereo vision, with a rate of 15 frames per second, and a resolution of 1024 x 768 pixels. Time of the internal clock of the computer recording the images is shown at the right top corner of the images, with a precision of seconds. The webcam was placed in front of the dVRK system so that PSM1 is on the right side of the images, and PSM2 on the left side. The camera projection matrix that relates world 3D points from the data files with their corresponding image projections is described in Section \ref{sec:calibration}.

\subsection{Data synchronization}
The kinematics data and the video data have been recorded using two different computers, both running on Ubuntu 16.04. The internal clocks of these computers have been synchronized in a common time reference using a Network Time Protocol (NTP) server. The internal clocks synchronization has been repeated before starting the experiment of a new user, if there was a break between users higher than one hour. 

As images and data have been recorded separately, although both computers are timely synchronized, video and data files does not start at the same time, i.e., for a particular trial, the timestamp of the first frame of the video does not correspond with the data of the first row of the corresponding data file. Thus, a manual synchronization between video and data files has been performed, in order to provide the initial video frame and data row with the same timestamp. For each trial, the synchronization procedure has been performed as follows:
\begin{enumerate}
    \item As the time shown in the videos has a precision of seconds, for each trial, we have manually search the frame with the first seconds break.
    \item Then, we have search the row in the corresponding data file with the timestamp corresponding with the time shown in the video, i.e., from the 50 samples for each seconds, we selected the row corresponding with the first one. Thus, the maximum error at this synchronization point is 20ms. 
\end{enumerate}

This manual synchronization is stored in the file \textit{'synchronizationData.txt'}, with the following structure: \textless trial\textgreater\ \textless initial\_frame\textgreater\ \textless initial\_row\textgreater. Thus, for using the data, video frames before the 'initial\_frame' and data corresponding with rows before the 'initial\_row' must be obviated. 

\subsection{Exercises evaluation}
The file 'scores.csv' contains the evaluation of each exercise according to the scoring of Table \ref{table:protocol}. Thus, for each trial, it features the execution time of the task (in seconds), the number of errors, and the final score. 

\subsection{Personal questionnaire}
After completing the experiment, participants were asked to fill-in a form to collect personal data that could be useful for further studies and analysis of the data. The form has questions related with the following items: age, dominant hand, task preferences, medical background, previous experience using the da Vinci or any teleoperated device, and hand-eye coordination skills. Questions demanding a level of expertise are multiple choice ranging from 1 (low) to 5 (high). 

\section{Camera calibration}\label{sec:calibration}
To facilitate the conversion between 3D real world coordinates from the data files to 2D image coordinates, the camera projection matrix for PSM1 ($C_{1}$) and PSM2 ($C_{2}$) have been computed. A projection matrix, $C$, is used to convert from a 3D real world point $P = [X,Y,Z]$ to the corresponding 2D image point $I = [u,v]$:
\begin{equation}
    \lambda 
    \begin{bmatrix}
        I \\1
    \end{bmatrix}
    = C
    \begin{bmatrix}
       P\\1
    \end{bmatrix}
\end{equation}

The projection matrixes for both tools have been computed by manual annotation of the tool tips in the image for 116 frames of six different trials. In particular, we have chosen the first two trials of each of the three tasks: pea on a peg task, post and sleeve,and wire chaser. The resulting projection matrixes are:

\begin{equation}
    C_{1} = 
    \begin{bmatrix}
          -42.1940&   15.1531&   13.2320&    6.7212\\
            2.4512&  -28.8705&   49.7470&    0.9284\\
           -0.0009&    0.0216&    0.0336&    0.0076\\
    \end{bmatrix}
\end{equation}

\begin{equation}
    C_{2} = 
    \begin{bmatrix}
          -53.5366&     2.8045&    30.6474&    -0.0011\\
            2.7140&   -31.0924&    52.6142&     1.5262\\
            0.0023&     0.0084&     0.0349&     0.0088\\
    \end{bmatrix}
\end{equation}

These projection matrixes may be used for performing automatic tools tracking in the image using the 3D tool tips coordinates provided by the data files. 
An example of this task is shown in the video of the Supplementary Material. 

\section{Using the data}
To facilitate the usage of the data, we provide MATLAB code that visualizes and replicates the motion of the four manipulators of the dVRK for a performance stored in the ROSMA dataset. The application allows loading a data file and generates a mat file with the data. Moreover, the data can be reproduced using the GUI buttons. The time slider shows the time progress of the reproduction, and at the top-right side, the current timestamp and data frame are also displayed. 

The GUI also gives the option of reproducing the data in ROS. If the checkbox \textit{ROS} is on, the joint configuration of each manipulator is sent through the corresponding ROS topics: /dvrk/\textless name\textgreater/set\_position\_goal\_joint. Thus, if the ROS package of the dVRK (dvrk-ros) is running, the real platform or the simulated one using the visualization tool RVIZ, the system will replicate the motion performed during the data file trial. The dVRK package can be downloaded from GitHub\footnote{\url{https://github.com/jhu-dvrk/dvrk-ros}}, as well as our MATLAB code \footnote{\url{https://github.com/SurgicalRoboticsUMA/rosma_dataset}}. This package includes the MATLAB GUI and a folder with all the  files required to launch the dVRK package with the configuration used during the data collection. 

\section{Summary}
The ROSMA dataset is a large surgical robotics collection of data using the da Vinci Research Kit. The authors provide a video showing the experimental setup used for collecting the data, along with a demonstration of the performance of the three exercises and the MATLAB GUI for visualizing the data\footnote{\url{https://www.youtube.com/watch?v=gEtWMc5EkiA&feature=youtu.be}}.
The main strength of our dataset versus the JIGSAWS one (\cite{Gao2014JHU-ISIModeling}), is the large quantity of data recorded (154 kinematic data, images, tasks evaluation and questionnaire) and the higher number of users (twelve instead of six).
This high amount of data could facilitate to advance in the field of artificial intelligence applied to the automation of tasks in surgical robotics, as well as surgical skills evaluation and gesture recognition.

\bibliographystyle{IEEEtran}
\bibliography{references.bib}

\end{document}